\documentclass[11pt]{article}
\usepackage{coling2018}
\usepackage{times}
\usepackage{color}
\usepackage{latexsym}
\usepackage{multirow}
\usepackage{graphicx} 
\usepackage{amssymb}
\usepackage{amsmath}
\usepackage{subfigure}
\usepackage{fancyhdr}

\makeatletter
\newcommand{\@BIBLABEL}{\@emptybiblabel}
\newcommand{\@emptybiblabel}[1]{}
\makeatother
\usepackage{hyperref}

\title{Dependent Gated Reading for Cloze-Style Question Answering}

\author{
	Reza Ghaeini, Xiaoli Z. Fern, Hamed Shahbazi, and Prasad Tadepalli \\
	Oregon State University, Corvallis, OR, USA \\
	\{ghaeinim, xfern, shahbazh, tadepall\}@eecs.oregonstate.edu
}

\date{}

\begin{document}
	\maketitle
	\begin{abstract}		
		We present a novel deep learning architecture to address the cloze-style question answering task. Existing approaches employ reading mechanisms that do not fully exploit the interdependency between the document and the query. In this paper, we propose a novel \emph{dependent gated reading} bidirectional GRU network (DGR) to efficiently model the relationship between the document and the query during encoding and decision making. Our evaluation shows that DGR obtains highly competitive performance on well-known machine comprehension benchmarks such as the Children's Book Test (CBT-NE and CBT-CN) and Who DiD What (WDW, Strict and Relaxed). Finally, we extensively analyze and validate our model by ablation and attention studies.
	\end{abstract}
	
	\section{Introduction}
Human language comprehension is an important and challenging task for machines that requires semantic understanding and reasoning over clues. The goal of this general task is to read and comprehend the given document and answer queries. 
	
	Recently, the cloze-style reading comprehension problem has received increasing attention from the NLP community. A cloze-style query \cite{cloze} is a short passage of text containing a blank part, which we must fill with an appropriate token based on the reading and understanding of a related document.
	The recent introduction of several large-scale datasets of cloze-style question answering made it feasible to train deep learning systems for such task \cite{wdw,cbt,cnn}. Various deep learning models have been proposed and achieved reasonable results for this task \cite{fg-reader,ga-reader,nse,aoa,epi-reader,as-reader,stanford-ar,cas,iterate-att}. The success of recent models are mostly due to two factors: 1) Attention mechanisms \cite{nmt}, which allow the model to sharpen its understanding and focus on important and appropriate subparts of the given context; 2) Multi-hop architectures, which read the document and/or the query in multiple passes, allowing the model to re-consider and refocus its understanding in later iterations.  Intuitively, both attention mechanisms and multi-hop reading fulfill the necessity of considering the dependency aspects of the given document and the query. Such a consideration enables the model to pay attention to the relevant information and ignore the irrelevant details.  Human language comprehension is often performed by jointly reading the document and query to leverage their dependencies and stay focused in reading and avoid losing relevant contextual information. Current state-of-the-art models also attempt to capture this by using the reading of the query to guide the reading of the document \cite{fg-reader,ga-reader}, or using the memory of the document to help interpret the query \cite{nse}. However, these systems only consider uni-directional dependencies. Our primary hypothesis is that we can gain further improvements 
by considering bidirectional dependencies.

	In this paper, we present a novel multi-hop neural network architecture, called Dependent Gated Reading (DGR), which addresses the aforementioned gap and performs dependent reading in both directions. Our model begins with an initial reading step that encodes the given query and document, followed by an iterative reading module (multi-hop) that employs soft attention to extract the most relevant information from the document and query encodings to augment each other's representation, which are then passed onto the next iteration of reading. Finally, the model performance a final round of attention allocate and aggregate to rank all possible candidates and make prediction.
	
	We evaluate our model on well-known machine comprehension benchmarks such as the Children's Book Test (CBT-NE \& CBT-CN), and Who DiD What (WDW, Strict \& Relaxed). Our experimental results indicate the effectiveness of DGR by achieving state-of-the-art results on CBT-NE, WDW-Strict, and WDW-Relaxed. In summary, our contributions are as follows: 1) we propose a new deep learning architecture to address the existing gap of reading dependencies between the document and the query. The proposed model outperforms the state-of-the-art for CBT-NE, WDW-Strict, and WDW-Relaxed by $0.5\%, 0.8\%$, and $0.3\%$ respectively; 2) we perform an ablation study and analysis to clarify the strengths and weaknesses of our model while enriching our understanding of the language comprehension task.
	
	\section{Related Work}

	The availability of large-scale datasets \cite{wdw,cbt,cnn} has enabled researchers to develop various deep learning-based architectures for language comprehension tasks such as cloze-style question answering.
	
	Sordoni et al. \shortcite{iterate-att} propose an \emph{Iterative Alternative Attention} (IAA) reader. IAA is a multi-hop comprehension model which uses a GRU network to search for correct answers from the given document. IAA is the first model that does not collapse the query into a single vector. It deploys an iterative alternating attention mechanism that collects evidence from both the document and the query. 
	
    Kadlec et al. \shortcite{as-reader} Introduce a single-hop model called \emph{Attention Sum Reader} (AS Reader) that uses two bi-directional GRUs (Bi-GRU) to independently encode the query and the document. It then computes a probability
distribution over all document tokens by
taking the softmax of the dot product between the query and the token representations. Finally, it introduces a \emph{pointer-sum attention} aggregation mechanism to aggregate the probability of multiple appearances of the same candidate. The candidate with the highest probability will be considered as the answer. Cui et al. \shortcite{aoa} introduce a similar single-hop model called \emph{attention-over-attention} (AOA) reader which uses a two-way
attention mechanism to allow the query and document
to mutually attend to one another.
	
	Trischler et al. introduce EpiReader \shortcite{epi-reader}, which uses AS Reader to first narrow down the candidates, then replaces the query placeholder with each candidate to yield a different query statement, and estimate the entailment between the document and the different query statements to predict the answer.
	
Munkhdalai and Yu \shortcite{nse} (NSE) propose a computational hypothesis testing framework based on memory augmented neural networks. They encode the document and query independently at the beginning and then re-encode the query
(but not the document) over multiple iterations (hops). At the end of each iteration, they predict an answer. The final answer is the candidate that obtains the highest probability over all iterations.

Dhingra et al. \shortcite{ga-reader} extend the AS Reader by proposing \emph{Gated Attention Reader} (GA Reader). GA Reader uses a multi-hop architecture to compute the representation of the documents and query. In each iteration the query is encoded independent of the document and previous iterations, but the document is encoded iterative considering the previous iteration as well as an attention mechanism with multiplicative gating to generate query-specific document representations. GA reader uses the same mechanism for making the final predictions as the AS reader. Yang et al. \shortcite{fg-reader} further extend the GA Reader with a fine grained gating approach that uses external semantic and syntactic features (i.e. NER, POS, etc) of the tokens to combine the word and character level embeddings and produce a final representation of the words. 		
	
	Among the aforementioned models, the GA Reader is the closest to our model in that we use a similar architecture that is multi-hop and performs iterative reading. The main distinct between our model and the GA Reader is the reading and encoding of the query. Instead of performing independent reading of query in each iteration, our reading and encoding of the query not only depends on the document but also the reading of previous iterations.

	Although cloze-style question answering task is well studied in the literature, the potential of dependent reading and interaction between the document and the query is not rigorously explored. In this paper, we address this gap by proposing a novel deep learning model (DGR). Experimental results demonstrate the effectiveness of our model.
	
\section{Dependent Gated Reading}
\begin{figure}[ht]
		\centering
		\includegraphics[width=\textwidth]{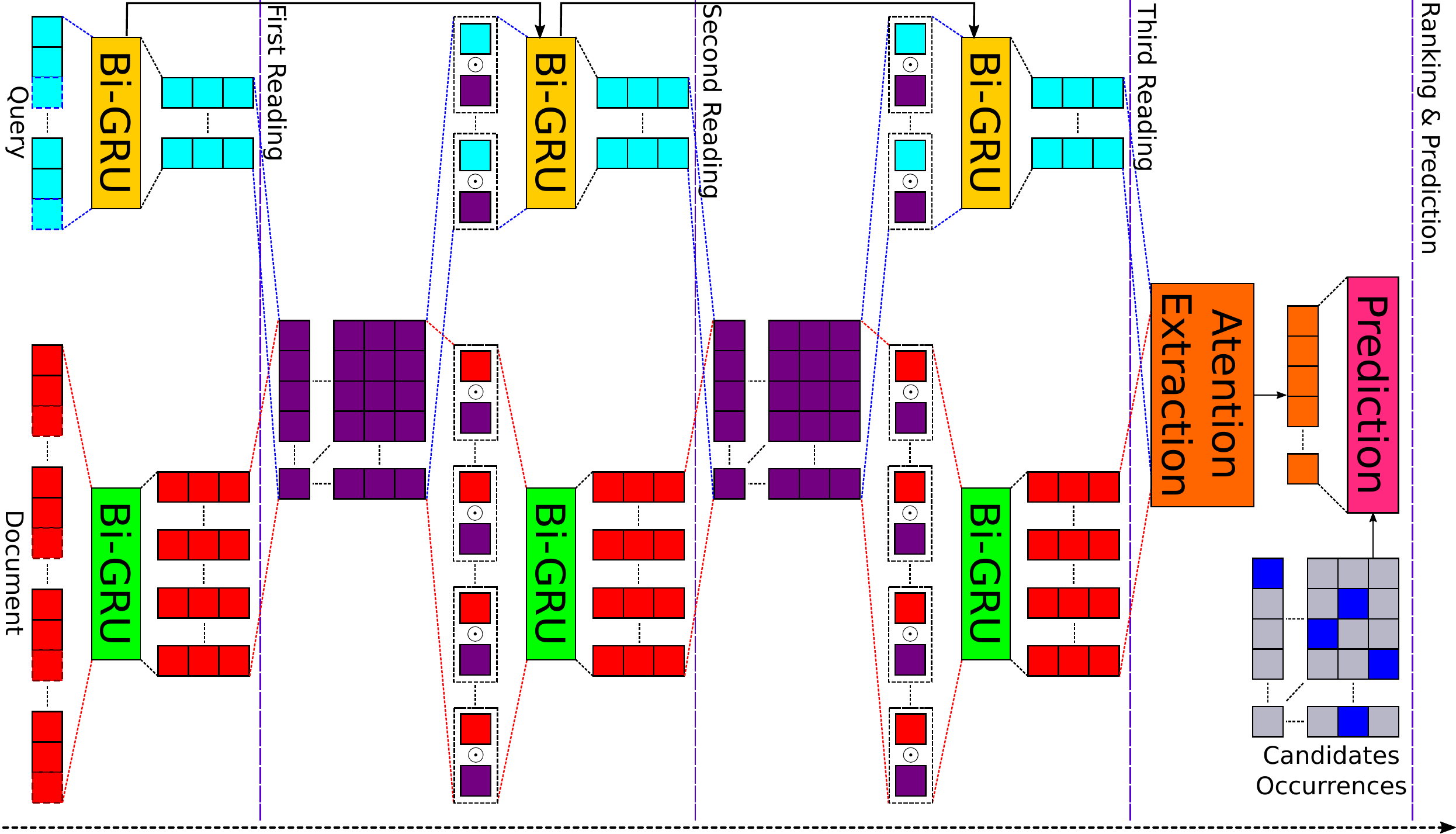}
		\caption{A high-level view of dependent gated reading model (DGR). The data (document $d$ and query $q$, depicted with red and cyan tensors respectively) flows from left to right. At the first (input) layer, the word representations are shown with black solid borders while the character representations are shown with colored dashed borders. The figure is color coded; relevant tensors and elements are shown with the same color. Note that none of the elements share parameters. The purple matrices extract relevant information between document and query representations. The black arrows between the query Bi-GRUs (yellow ones) pass the final hidden state of a Bi-GRU to another one as initialization value for its hidden state. \label{fig:model}}
	\end{figure}

	Figure~\ref{fig:model} depicts a high-level view of our proposed Dependent Gate Reading (DGR) model, which follows a fairly standard multi-hop architecture, simulating the multi-step reading and comprehension process of humans. 

	The input to our model at the training stage can be represented as a tuple $(D, Q, C, a)$, where $D=[d_1, \cdots, d_n]$ is the document of length $n$, $Q=[q_1, \cdots, q_m]$ is the query of length $m$ with a placeholder, $C=[c_1, \cdots, c_g]$ is a set of $g$ candidates and $a \in C$ is the ground truth answer. Here we assume $d_i, q_j$ are some form of embedding of the individual tokens of document and query. At the testing stage, given the input document $D$, query $Q$ and candidate set $C$, the goal is to choose the correct candidate $a$ among $C$ for the placeholder in $Q$.

	DGR can be divided to two major parts: Multi-hop Reading, and Ranking \& Prediction.

	\subsection{Multi-hop Reading of Document and Query}
	\label{sec:enc}
Recurrent networks provide a natural solution for modeling variable length sequences. Consequently, we use bi-directional Gated Recurrent Units (Bi-GRUs) \cite{gru} as the main building blocks for encoding the given document and query. For the initial step of our multi-hop reading, the document $D$ and the query $q$ are read with two separate Bi-GRUs (Equations~\ref{eq:encd0} and \ref{eq:encq0}) where $\hat{d}^0 \in \mathbb{R}^{n \times r}$ and
$\hat{q}^0 \in \mathbb{R}^{m \times r}$	are the first Bi-GRU reading sequences of $D$ and $Q$ respectively. $h^0$ consists of two parts, $h^0_f$ and $h^0_b$, which record the final output of forward and backward GRU reading of $Q$ respectively. Note that ``$-$'' in  equations means that we do not care about the associated variable and its value. 
	\begin{equation}
	\hat{d}^0, - = BiGRU_{d0}(D, 0)
	\label{eq:encd0}
	\end{equation}
	\begin{equation}
	\hat{q}^0, h^0 = BiGRU_{q0}(Q, 0)
	\label{eq:encq0}
	\end{equation} 
 
    We use $s\in [0,S]$ to denote the reading iteration, with $S+1$ total iterations. For the initial iteration ($s=0$),  both Bi-GRUs are fed with a zero vector for the initial hidden state as shown in Equations~\ref{eq:encd0} and ~\ref{eq:encq0}.     
Once the document and query encodings ($\hat{d}^{s}$ and $\hat{q}^{s}$ respectively) are computed, we employ a soft alignment method to associate the relevant sub-components between the given document and query. In deep learning models, this is often achieved with a soft attention mechanism. We follow the same soft attention mechanism as used in the GA reader~\cite{ga-reader}, which is described below for completeness. 

Given $\hat{d}^{s}$ and $\hat{q}^{s}$, we first compute the unnormalized attention weights between the $i$-th token of the document and the $j$-th token of the query as the similarity between the corresponding hidden states with Equation~\ref{eq:energy} (energy function).
	
	\begin{equation}
	e^s_{ij} = (\hat{d}^s_i)^T \hat{q}^s_j,  \quad  \forall i \in [1,n], \forall j \in [1,m], \forall s \in [0, S-1]
	\label{eq:energy}
	\end{equation}
	
    For each document token and query token, the most relevant semantics from the other context are extracted and composed based on $e^s_{ij} \in \mathbb{R}^{n \times m}$. Equations \ref{eq:att:d} and \ref{eq:att:q} provide the specific details of this procedure where $\tilde{d}^s_i  \in \mathbb{R}^{r}$ represents the extracted  information from the current reading of the query, $\hat{q}^s$, that is most relevant to the $i$-th document token by attending to $\hat{d}^s_i$.  Similarly $\tilde{q}^s_j  \in \mathbb{R}^{r}$ represents, for the $j$-th query token, the extracted relevant document information from $\hat{d}^s$ by attending to $\hat{q}^s_j$.
	
	\begin{equation}
	\tilde{d}^s_i = \sum_{j=1}^{m} \frac{\exp(e^s_{ij})}{\sum_{k=1}^{m} \exp(e^s_{ik})} \hat{q}^s_j, \quad  \forall i \in [1,n], \forall s \in [0, S-1]
	\label{eq:att:d}
	\end{equation}
	
	\begin{equation}
	\tilde{q}^s_j = \sum_{i=1}^{n} \frac{\exp(e^s_{ij})}{\sum_{k=1}^{n} \exp(e^s_{kj})} \hat{d}^s_i, \quad \forall j \in [1,m], \forall s \in [0, S-1]
	\label{eq:att:q}
	\end{equation}

To incorporate the context information, we use element-wise product of the tuples $(\hat{d}^s_i, \tilde{d}^s_i)$ or $(\hat{q}^s_j, \tilde{q}^s_j)$ to produce a new representation of the hidden states for the document and the query as described in Equations \ref{eq:epd} and \ref{eq:epq}.

	\begin{equation}
	u^s_i = \hat{d}^s_i \odot \tilde{d}^s_i, \quad \forall s \in [0,S-1]
	\label{eq:epd}
	\end{equation}
	\begin{equation}
	v^s_j = \hat{q}^s_j \odot \tilde{q}^s_j, \quad \forall s \in [0,S-1]
	\label{eq:epq}
	\end{equation}
	
	Here $\odot$ stands for element-wise product, and $u^s \in \mathbb{R}^{r}$ and $v^s \in \mathbb{R}^{r}$ are the new encodings of the document and query respectively.Note that GA-reader uses the same mechanism to update the document encoding but does not change the query representation according to the document.

    We then pass the new document ($u^s$) and query ($v^s$) embeddings to the Bi-GRUs for the next iteration $s+1$. Note that for query reading, we feed, $h^{s}$, the final hidden state of the previous reading (without document based updates) to the Bi-GRU of the next iteration as the initial hidden state. 
Intuitively, $h^{s}$ provides a summary understanding of the query from the previous iteration, without the document modulated updates. By considering both $h^{s}$ and $v^{s}$, this encoding mechanism provides a richer representation of the query. This is formally described by Equations \ref{eq:encds} and \ref{eq:encqs}. 
	\begin{equation}
	\hat{d}^{s+1}, - = BiGRU_{ds}(u^{s}, 0), \forall s \in [0,S-1]\
	\label{eq:encds}
	\end{equation}
	\begin{equation}
	\hat{q}^{s+1}, h^{s+1} = BiGRU_{qs}(v^{s}, h^{s}), \forall s \in [0,S-1]
	\label{eq:encqs}
	\end{equation} 	
	
We should note that using the following configuration variations did not yield any improvement to our model: 1) Other choices for gating aggregation strategy (Equations~\ref{eq:epd} and \ref{eq:epq}) like addition, concatenation, or applying a transformation function on different sub-members of $\{$element-wise product, concatenation and difference$\}$. 2) Residual connection. 
	
\subsection{Ranking \& Prediction}
Given the final document and query encodings, $\hat{d}^{S}$ and $\hat{q}^{S}$, the final stage of our model computes a score for each candidate $c\in C$. This part of our model use the same \textit{point sum attention} aggregation operation as introduced by the Attention Sum (AS) reader~\cite{as-reader}, which is also used by the GA reader~\cite{ga-reader}.

Let $idx$ be the position of the the placeholder in $Q$, and $\hat{q}_{idx}^{S}$ be the associated hidden embedding of the placeholder in the given query. We first compute the probability of each token in the document to be the desired answer by computing the dot product between $\hat{q}_{idx}^{S}$ and $\hat{d}_j^{S}$ for $j=1,...,n$ and then normalize with the softmax function:  

	\begin{equation}
	y = \text{softmax}((\hat{q}^{S}_{idx})^T \hat{d}^{S})
	\label{eq:pc}
	\end{equation}

	\noindent where $y \in \mathbb{R}^{n}$ gives us a normalized attention/probability over all tokens of the document. Next, the probability of each particular candidate $c \in C$ for being the answer is computed by aggregating the document-level attentions of all  positions in which $c$ appears: 
	
	\begin{equation}
	p(c|D,Q) \propto \sum_{i \in I(c,D)} y_i, \quad \forall c \in C
	\label{eq:p}
	\end{equation}
	\noindent where $I(c, D)$ indicates the positions that candidate $c$ appears in the document $D$ (Candidate Occurrences in Figure~\ref{fig:model}). Finally the prediction is given by $a^* = \textit{argmax}_{c \in C} \textit{ } p(c|D,Q)$.

\paragraph{Key differences from the GA reader.}
Given the strong similarity between our model and the GA reader, it is worth highlighting the three key differences between the two models: (a) Document gated query reading: we compute a document-specific query representations to pass to the next query reading step; (b) Dependent query reading: in each iteration, the input to the query BiGRU comes from the document gated encoding of the query from the last iteration whereas the GA Reader reads the queries independently in all iterations; (c) Dependent query BiGRU initialization: the query BiGRU is initialized with the final hidden states of the query BiGRU from the previous iteration.
These key differences in query encoding are designed to better capture the interdependences between query and document and produce richer and more relevant representations of the query and enhance the comprehension and query answering performance.

\subsection{Further Enhancements}
\label{sec:enh}
Following the practice of GA reader, we included several enhancements which have been shown to be helpful in previous work. 
\paragraph{Question Evidence Common Word Feature.}  To generate the final document encoding $\hat{d}^{S}$, an additional modification of $u^{S-1}$ is introduced before applying Equation~\ref{eq:encds}. 
Specifically, an additional \emph{Question Evidence Common Word Feature} (qe-comm) \cite{feat} is introduced for each document token, indicating whether the token is present in the query. Assume $f_i$ stands for the qe-comm feature of the $i$-th document token, therefore, $u^{S-1}_i = [u^{S-1}_i, f_i]$. 
\paragraph{Character-level embeddings.}	Word-level embeddings are good at representing the semantics of the tokens but suffers from out-of-vocabulary (OOV) words and is incapable of representing sub-word morphologies. Character-level embeddings effectively address such limitations \cite{charlvl,t2v}. In this work, we represent a token by concatenating its word embedding and character embedding.  
To compute the character embedding of a token $w=[x_1, \cdots, x_l]$, we pass $w$ to two GRUs in forward and backward directions respectively. Their outputs are then concatenated and passed through a linear transformation to form the character embedding of the token. 

\section{Experiments and Evaluation}
\subsection{Datasets} \label{sec:data}
	We evaluate the DGR model on three large-scale language comprehension datasets, Children's Book Test Named Entity (CBT-NE), Common Noun (CBT-CN), and Who Did What (WDW) Strict and Relaxed.

	The first two datasets are formed from two subsets of the Children's Book Test (CBT) \cite{cbt}. Documents in CBT consist of 20 contiguous sentences from the body of a popular children's book, and queries are formed by replacing a token from the $21^{st}$ sentence with a placeholder. We experiment on subsets where the replaced token is either a named entity (CBT-NE) or common noun (CBT-CN). Other subsets of CBT have also been studied previously but because simple language models have been able to achieve human-level performance on them, we ignore such subsets \cite{cbt}.
	
	The Who Did What (WDW) dataset \cite{wdw} is constructed from the LDC English Gigaword newswire corpus. Each sample in WDW is formed from two independent articles. One article is considered as the passage to be read and the other article on the same subject is used to form the query. Missing tokens are always person named entities. For this dataset, samples that are easily answered by simple systems are filtered out, which makes the task more challenging. There are two versions for the training set (Strict and Relaxed) while using the same development and test sets. Strict is a small but focused/clean training set while Relaxed is a larger but more noisy training set. We experiment on both of these training sets and report corresponding results on both settings. Statistics of all the aforementioned datasets are summarized in Table \ref{tab:data:stat} of the appendix.
	
Other datasets for this task include CNN and Daily Mail News \cite{cnn}. Because previous models already achieved human-level performance on these datasets, following Munkhdalai and Yu \shortcite{nse}, 
we do not include them in our study.
	
\subsection{Training Details \& Experimental Setup}
We use pre-trained $100$-$D$ Glove $6B$ vectors ~\cite{glove} to initialize our word embeddings while randomly initializing the character embedding. All hidden states of BiGRUs have $128$ dimensions ($o=100$ and $r=128$). The weights are learned by minimizing the negative log-loss (Equation~\ref{eq:cost}) on the training data via the Adam optimizer~\cite{adam}. The learning rate is 0.0005. To avoid overfitting, we use dropout~\cite{dropout} with rate of 0.4 and 0.3 for CBT and WDW respectively as regularization, which is applied to all feedforward connections. While we fix the word embedding, character embeddings are updated during the training to learn effective representations for this task. We use a fairly small batch size of 32 to provide more exploration power to the model. 
	\begin{equation}
		L = \sum_i - \log(p(a|D,Q))
		\label{eq:cost}
	\end{equation}
	
	\subsection{Results}
	
	Table~\ref{tab:result} shows the test accuracy of the models on CBT-NE, CBT-CN, WDW-Strict, and WDW-Relaxed. We divide the previous models into four categories: 1) Single models (rows 1-5), 2) Ensemble models (rows 6-9), 3) NSE models (rows 10-14), and 4) the FG model (row 15). Table~\ref{tab:result} primarily focuses on comparing models that do not rely on any NLP toolkit features (i.e. POS, NER, etc), with the exception of the FG model which uses additional information about document tokens including POS, NER and word frequency information to produce the embedding of the token.
	
	\begin{table}[t!]
		\small
		\begin{center}
			\begin{tabular}{lcccc}
				\hline
				\multirow{2}{*}{\textbf{Method}} & \multicolumn{4}{c}{\textbf{Test Accuracy(\%)}} \\ \cline{2-5} 
				& CBT-NE & CBT-CN & WDW-Strict & WDW-Relaxed \\ \hline
				AS Reader~\cite{as-reader} & 68.6\% & 63.4\% & 57.0\% & 59.0\% \\
				EpiReader~\cite{epi-reader} & 69.7\% & 67.4\% & - & - \\
				IAA Reader~\cite{iterate-att} & 68.6\% & 69.2\% & - & - \\
				AOA Reader~\cite{aoa} & 72.0\% & 69.4\% & - & - \\
				GA Reader~\cite{ga-reader} & 74.9\% & 70.7\% & 71.2\% & 72.6\% \\
				\hline
				AS Reader (Ensemble)~\cite{as-reader} & 70.6\% & 68.9\% & - & - \\
				EpiReader (Ensemble)~\cite{epi-reader} & 71.8\% & 70.6\% & - & - \\
				IAA Reader (Ensemble)~\cite{iterate-att} & 72.0\% & 71.0\% & - & - \\
				AOA Reader (Ensemble)~\cite{aoa} & 74.5\% & 70.8\% & - & - \\
				\hline
				NSE (T=1)~\cite{nse} & 71.1\% & 69.7\% & 65.5\% & 65.3\% \\
				NSE Query Gating (T=2)~\cite{nse} & 71.5\% & 70.7\% & 65.1\% & 65.5\% \\
				NSE Query Gating (T=6)~\cite{nse} & 71.4\% & 72.0\% & 65.7\% & 65.8\% \\
				NSE Adaptive Computation (T=2)~\cite{nse} & 72.1\% & 71.2\% & 65.4\% & 66.0\% \\
				NSE Adaptive Computation (T=12)~\cite{nse} & 73.2\% & 71.4\% & 66.2\% & 66.7\% \\
				\hline
				FG~\cite{fg-reader} & 74.9\% & 72.0\% & 71.7\% & 72.6\% \\
				\hline
				DGR & \textbf{75.4\%} & 70.7\% & \textbf{72.0\%} & \textbf{72.9\%} \\
				\hline
			\end{tabular}
		\end{center}
		\caption{\label{tab:result} Performance of proposed model (DGR) on the test set of CBT-NE, CBT-CN, WDW-Strict, and WDW-Relaxed datasets.}
	\end{table}
	 
From Table~\ref{tab:result}, we can see that DGR achieves the state-of-the-art results on all aforementioned datasets expect for CBT-CN. The targets of CBT-NE, WDW-Strict, and WDW-Relaxed are all Named Entities while the CBT-CN focuses on Common Noun. We believe that our architecture is more suitable for Named Entity targeted comprehension tasks. This phenomenon warrants a closer look in future work.
	Comparing GA Reader, FG, and DGR (the three models with similar architectures), we see that FG outperform the GA Reader on CBT-CN and WDW-Strict datasets while DGR outperforms both FG and GA Reader results on CBT-NE, WDW-Strict, WDW-Relaxed datasets with noticeable margins. This suggests that while the NLP toolkit features such as POS and NER could help the performance of the comprehension models (specially in CBT-CN), capturing richer dependency interaction between document and query appears to play a more important role for comprehension tasks focusing on Named Entities.
	
	Finally, For each of the three datasets on which our model achieves the state-of-the-art performance, we conducted the one-sided McNemar’s test to verify the statistical significance of the performance improvement over the main competitor (GA reader). The obtained p-values are 0.03, 0.003, and 0.011 for CBT-NE, WDW-Strict, and WDW-Relaxed respectively, indicating that the performance gain by DGR is statistically significant.
	
	\subsection{Ablation Study}
	We conducted an ablation study on our model to examine the importance and the effect of proposed strategies. We investigate all settings on the development set of the CTB-NE, CBT-CN, WDW-Strict, and WDW-Relaxed datasets. Consider the three key differences of our method from the GA Reader: (a) Document gated query reading --- here we compute a document-specific query representations to pass to the next reading layer; (b) Dependent query reading --- the query readings are dependent from one layer to the next as the input to the next reading layer comes from the output of previous layer; (c) Dependent BiGRU initialization --- query BiGRUs of a later layer are initialized with the final hidden states of previous layer's query BiGRU.

	\begin{table}[t!]
		\small
		\begin{center}
			\begin{tabular}{lcccc}
				\hline
				\multirow{2}{*}{\textbf{Method}} & \multicolumn{4}{c}{\textbf{Development Accuracy(\%)}} \\ \cline{2-5} 
				& CBT-NE & CBT-CN & WDW-Strict & WDW-Relaxed \\ \hline
				1) DGR & \textbf{77.90} & \textbf{73.80} & \textbf{71.78} & \textbf{72.26} \\ \hline
				2) DGR - (a) & 75.60 & 72.25 & 71.04 & 71.82 \\
				3) DGR - (c) & 77.50 & 72.45 & 71.29 & 71.93 \\ \hline
				4) DGR - (a) \& (b)  & 77.85 & 73.05 & 71.67 & 72.20 \\
				5) DGR - (a) \& (c) & 76.00 & 72.85 & 71.37 & 72.13 \\ \hline
				6) DGR - (a) \& (b) \& (c) & 77.65 & 73.00 & 71.61 & 72.16 \\
				\hline
			\end{tabular}
		\end{center}
		\caption{\label{tab:ablation} Ablation study results. Performance of different configurations of the proposed model on the development set of the CBT-NE, CBT-CN, WDW-Strict, and WDW-Relaxed datasets}
	\end{table}
	
	Table~\ref{tab:ablation} shows the ablation study results on the development set of CBT-NE, CBT-CN, WDW-Strict, and WDW-Relaxed for a variety of DGR configurations by removing one or more of the key differences with GA reader. Note that by all removing all three difference elements, configuration 6 reduces to the GA reader. 
	
According to Table~\ref{tab:ablation}, DGR achieves the best development accuracy on all datasets which indicates that collectively, the three elements lead to improved effectiveness. 

\paragraph{Effect of document dependent reading.} Configuration 2 removes the document dependent reading, and retains the other two elements. Interestingly, this configuration achieved the worst performance among all variations. Without proper guiding from the document side, iteratively reading the query actually leads to worse performance than independent query reading. This suggests that document dependent reading is a critical element that helps achieve better reading of query. 

\paragraph{Effect of Dependent Query BiGRU initialization.} 
In Configuration 3, we remove the dependent query BiGRU initialization, which results in a performance loss ranging from 0.33\% (WDW-relaxed) to 1.35\% (CBT-CN), suggesting that this connection provides important information that helps the reading of the query. Note that simply adding dependent query BiGRU initialization to GA reader (configuration 4) leads to a slight improvement over GA reader, which again confirms the usefulness of this channel of information. 

\paragraph{Effect of dependent query reading.}
Unfortunately, we cannot only remove (b) from our model because it will cause dimension mismatch between the document and query representation preventing the gating operation for computing the document gated query representation. Instead, we compare the GA reader (configuration 6) with configure 5, which adds dependent query reading to the GA reader. We can see that adding the dependent query reading to the GA reader actually leads to a slight performance loss. Note that further including document gated reading (configuration 3) improves the performance on CBT-NE, but still fails to outperform GA reader. This points to a potential direction to further improve our model by designing a new mechanism that is capable of document dependent gating without the dependent query reading. 
	
	\subsection{Analysis} \label{sec:err}
    
	In this section, We first investigate the performance of DGR and its variations on two attributes: the \emph{document length}, and \emph{query length}. Then we show a layer-wise visualization of the energy function (Equation~\ref{eq:energy}) for an instance from the CBT-NE dataset.
	
\subsubsection{Length Study}
	
    Among the four datasets that we use in this paper, WDW-Relaxed is the biggest and the most noisy one which makes it as a good candidate for analyzing the trend and behavior of our models.
	
	\begin{figure}[ht]
		\centering
		\includegraphics[width=.9\textwidth]{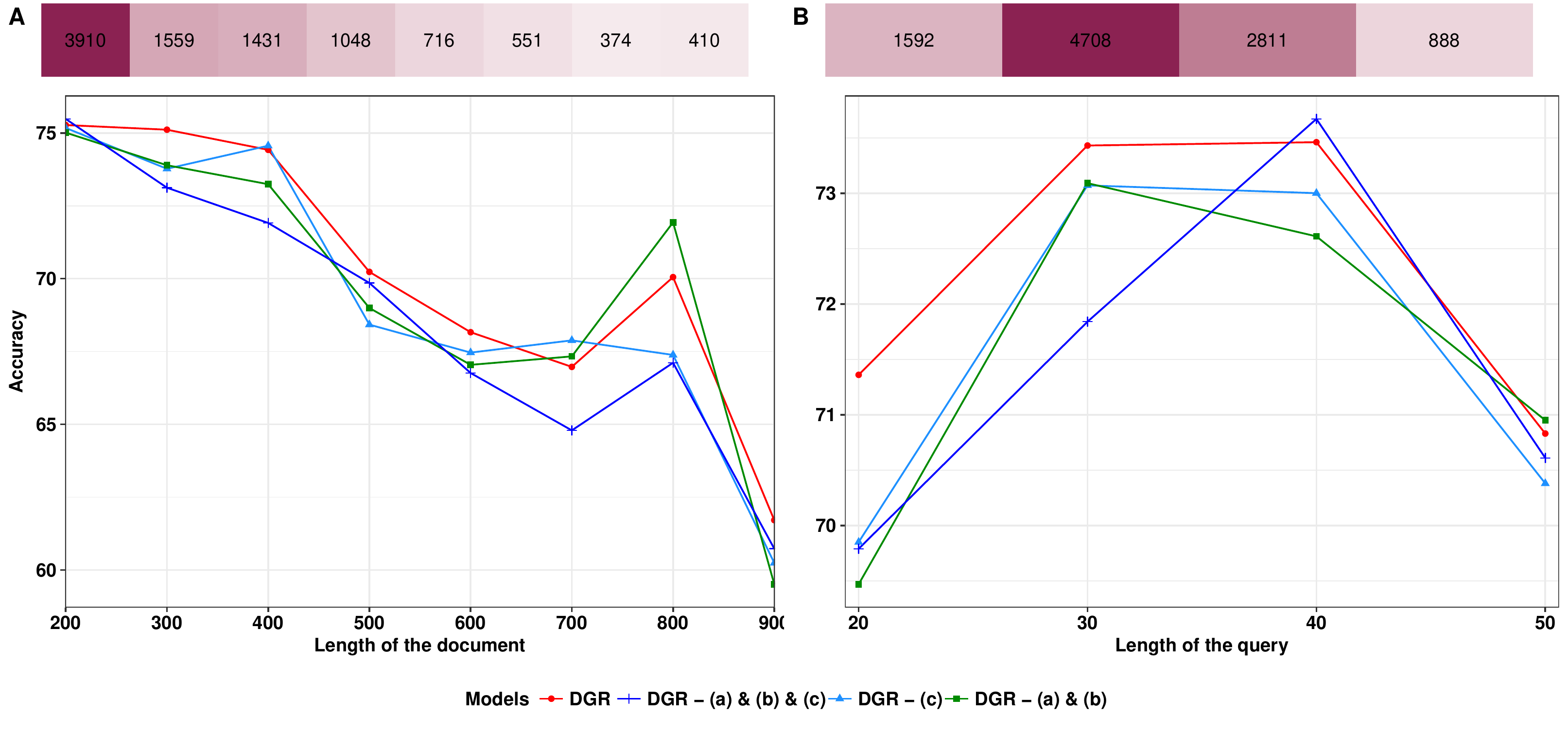}
		\caption{Test accuracy of DGR and its variations against the length of the document (A), and  length of the query (B) on the WDW-Relaxed dataset. The bar on top of each figure indicates the number of samples in each interval. Darker color in the bars illustrates more samples. \label{fig:line_plot}}
	\end{figure}
    
    Figure~\ref{fig:line_plot} depicts the performance of DGR and its variations against the length of document (left), and the length of query (right). A bar on top of each diagram indicates the frequency of samples in each intervals. Each data sample is added to the closet interval. 
    
    Overall Figure~\ref{fig:line_plot} suggests that DGR  achieves highly competitive performance across different document and query lengths in comparison to the other variations including the GA reader. In particular, DGR perform better or similarly to the GA reader (``DGR - (a) \& (b) \& (c)'') in all categories except when query length is between 30 and 40 where GA reader wins with a small margin. Furthermore, we see that ``DGR - (a) \& (b)'' wins over ``DGR - (a) \& (b) \& (c)'' in most document length categories. This suggests the positive effect of the connection offered by (c), especially for longer documents.
    
\subsubsection{Attention Study}
    To gain insights into the influence of the proposed strategies on the internal behavior of the model, we analyze the attention distribution at intermediate layers. We show a visualization of layer-wise normalized aggregated attention weights (energy function, Equation~\ref{eq:energy}) for candidate set over the query (for more examples look at Section \ref{sec:apx:att} of the appendix). In each figure, the top plots show the layer-wise attention of DGR and the bottom plots show the layer-wise attention of the GA reader, i.e., ``DGR - (a) \& (b) \& (c)''. Moreover, the left and middle plot show the aggregated attention of candidates over the whole query while the right plot depicts the aggregated attention of the candidates for the placeholder in the query in  the final layer.
    
   \begin{figure}[ht]
	\centering
		\includegraphics[width=\textwidth]{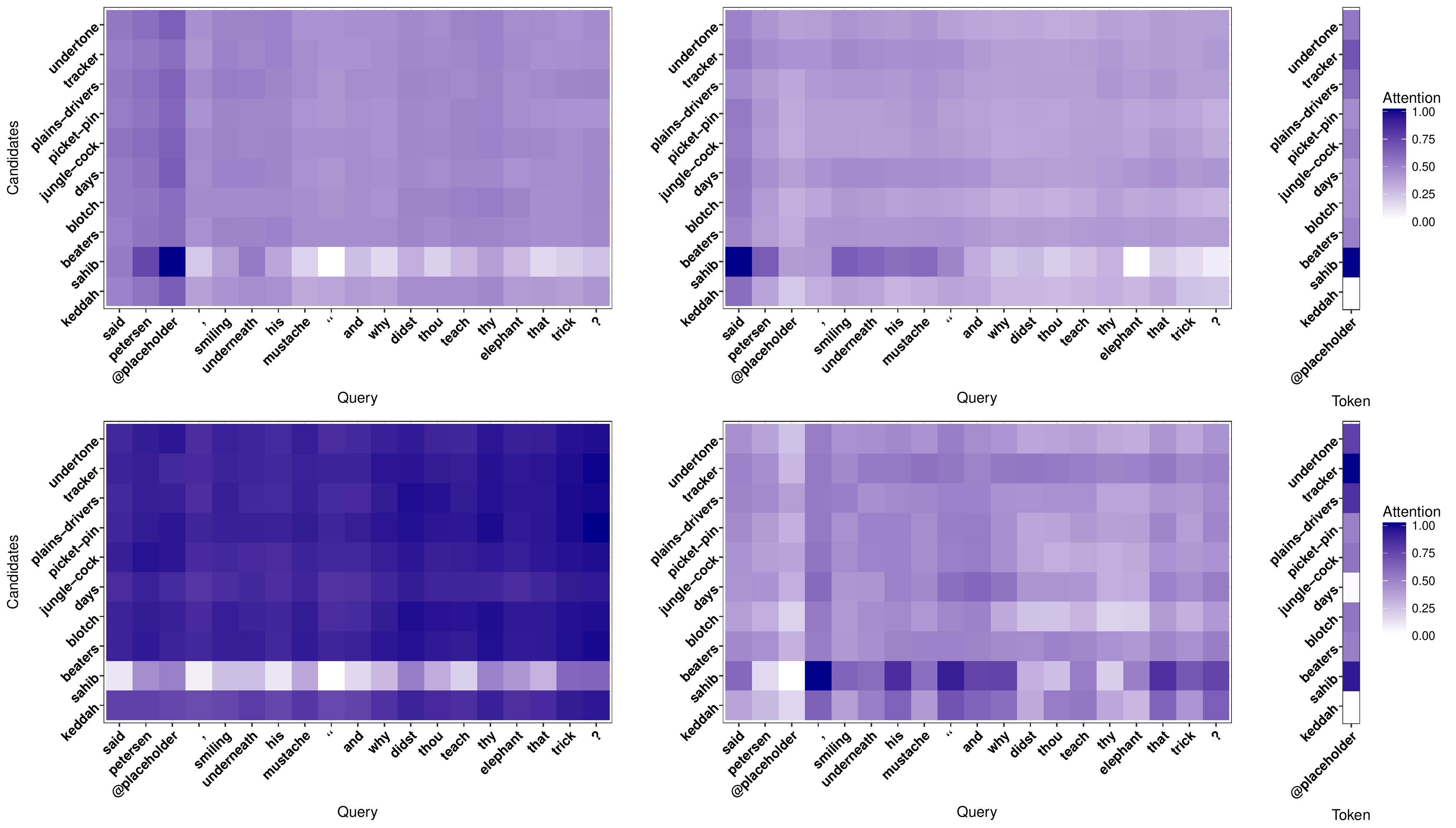}
		\caption{Layer-wise normalized attention visualization of ``DGR'' (top) and ``DGR - (a) \& (b) \& (c)'' (bottom) for a sample from the CBT-NE test set. Darker color illustrates higher attention. Figures only show the aggregated attention of candidates. The gold answer is ``sahib''. \label{fig:attention}}
	\end{figure}
    
A generic pattern observed in our study is that GA reader tends to generate more uniform attention distributions while DGR produces more focused attention. In other words, each layer of DGR tends to focus on different sub-parts and examine different hypotheses, illustrating the significant impact of the proposed strategies on the attention mechanism.

\section{Conclusion}
    
    We proposed a novel cloze-style question answering model (DGR) that efficiently model the relationship between the document and the query. Our model achieves the the state-of-the-art results on several large-scale benchmark datasets such as CBT-NE, WDW-Strict, and WDW-Relaxed. Our extensive analysis and ablation studies confirm our hypothesis that using a more sophisticated method for modeling the interaction between document and query could yield further improvements.
   	
	\bibliographystyle{acl}
	\bibliography{acl2018}
	
	\appendix
	
	\section{Dataset Statistics}
	\label{sec:apx:data}

	\begin{table}[ht]
		\small
		\begin{center}
			\begin{tabular}{lcccc}
				\hline
				& CBT-NE & CBT-CN &  WDW-Strict & WDW-Relaxed \\ \hline\hline
				\# training set & 108,719 & 120,769 & 127,786 & 185,978 \\
				\# development set & 2,000 & 2,000 & 10,000 & 10,000 \\
				\# test set & 2,500 & 2,500 & 10,000 & 10,000 \\
				\# vocabulary & 53,063 & 53,185 & 347,406 & 308,02  \\
				max document length & 1,338 & 1,338 & 3,085 & 3,085 \\
				\hline
			\end{tabular}
		\end{center}
		\caption{\label{tab:data:stat} Dataset statistics}
	\end{table}

	\section{Rule-based Disambiguation Study}
	\label{sec:apx:rule}
	
	In this section, we present a simple rule-based detection strategy for CBT-NE dataset which disambiguates about $30\%$ and $18\%$ of the samples in CBT-NE development and test sets. For each query $q$, assume $w$ is previous/next next word in the placeholder which start with upper case character. If such a $w$ exists, we look for $w$ in the document $d$ and collect all words that could appears next/before $w$. After removing all collected words that are not in the candidate list $C$, the samples is disambiguated and solved if we end up with a single word (answer). We refer to the set of such samples as disambiguated set. Table~\ref{tab:rule:analyse} shows the statistics of this rule-based strategy on the rule-based disambiguated test set of CBT-NE. Furthermore, Table~\ref{tab:rule:sample} shows a data sample in CBT-NE that is correctly disambiguated with our rule-based approach.
	
	\begin{table}[ht]
		\small
		\begin{center}
			\begin{tabular}{lccc}
				\hline 
				Set & Correct Disambiguation(\%) & Wrong Disambiguation(\%) & Total Disambiguation \\ \hline
				Development & 29.65\% & 0.1\% & 595  \\ 
				Test & 18.36\% & 0.12\% & 462 \\
				\hline
			\end{tabular}
		\end{center}
		\caption{\label{tab:rule:analyse} Statistics and performance of the proposed rule-based strategy on CBT-NE dataset.}
	\end{table}
	
	\begin{figure}[ht]
		\centering
		\includegraphics[width=0.6\textwidth]{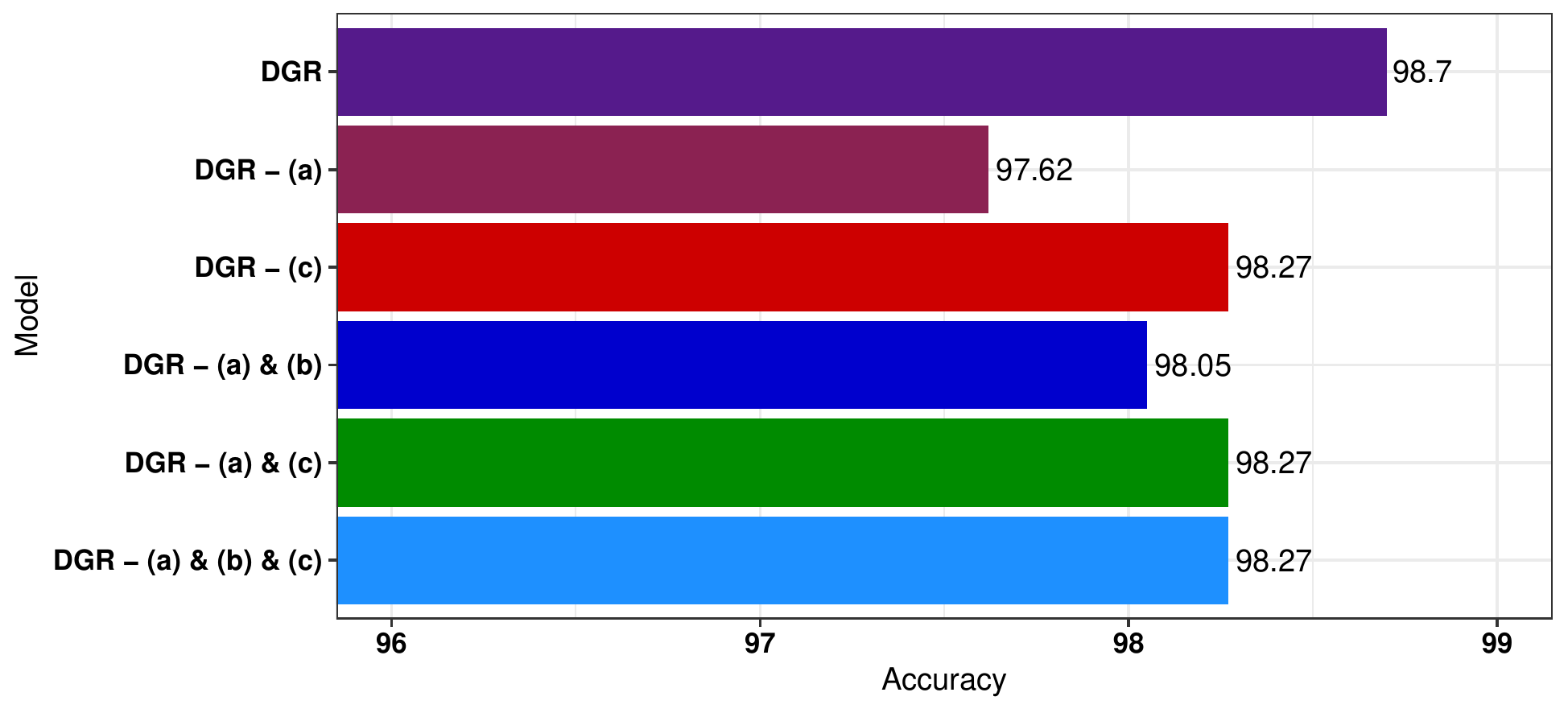}
		\caption{Performance of DGR and its variations on the rule-based disambiguated test set of CBT-NE.\label{fig:rule_barl}}
	\end{figure}
	
	Figure~\ref{fig:rule_barl} shows the performance of DGR and its variations on the set of data samples in CBT-NE test set that could be disambiguated with the proposed rule-based strategy. Although we use the lower case words in the training process, all models perform substantially well on disambiguating such samples. This observation could demonstrate the effectiveness of the general architecture.
	
	\begin{table}[t!]
		\begin{center}
			\begin{tabular}{ll}
				\hline
				doc$^a$ &  1 Instead of answering , \textbf{Jimmy} {\color{blue} Skunk} began to laugh . \\
				& 2 `` Who 's a bug ? '' \\
				& 3 demanded Old Mr. Toad , more crossly than before . \\
				& 4 `` There is n't any bug , Mr. Toad , and I beg your pardon , \\
				& '' replied \textbf{Jimmy} , remembering his politeness . \\
				& 5 `` I just thought there was . \\
				& 6 You see , I did n't know you were under that piece of bark . \\
				& 7 I hope you will excuse me , Mr. Toad . \\
				& 8 Have you seen any fat beetles this morning ? '' \\
				& 9 `` No , '' said Old Mr. Toad grumpily , and yawned and rubbed his eyes . \\
				& 10 `` Why , '' exclaimed \textbf{Jimmy} {\color{blue} Skunk} , `` I believe you have just waked up ! '' \\
				& 11 `` What if I have ? '' \\
				& 12 demanded Old Mr. Toad . \\
				& 13 `` Oh , nothing , nothing at all , Mr. Toad , '' replied \textbf{Jimmy} {\color{blue} Skunk} , `` \\
				& only you are the second one I 've met this morning who had just waked up . '' \\
				& 14 `` Who was the other ? '' \\
				& 15 asked Old Mr. Toad . \\
				& 16 `` Mr. Blacksnake , '' replied \textbf{Jimmy} . \\
				& 17 `` He inquired for you . '' \\
				& 18 Old Mr. Toad turned quite pale . \\
				& 19 `` I -- I think I 'll be moving along , '' said he . \\
				& 20 XVII OLD MR. TOAD 'S MISTAKE If is a very little word to look at , \\
				& but the biggest word you have ever seen does n't begin to have so much  \\
				& meaning as little `` if . '' \\
				\hline
				query & 21 If \textbf{Jimmy} {\color{red} @placeholder} had n't ambled down the Crooked Little Path just \\
				& when he did ; if he had n't been looking for fat beetles ; if he had n't seen \\
				& that big piece of bark at one side and decided to pull it over ; if it had n't \\ 
				& been for all these `` ifs , '' why Old Mr. Toad would n't have made the \\
				& mistake he did , and you would n't have had this story .\\
				\hline
				cands$^b$ & Blacksnake, Jimmy, Mr., {\color{blue} Skunk}, Toad, XVII, bug, morning, pardon, second \\
				\hline
				ans$^c$ & {\color{blue} Skunk}\\
				\hline
				pred$^d$ & {\color{blue} Skunk}\\
				\hline
				\multicolumn{2}{l}{$^a$ doc, Document} \\
				\multicolumn{2}{l}{$^b$ cands, Candidates} \\
				\multicolumn{2}{l}{$^c$ ans, Answer} \\
				\multicolumn{2}{l}{$^d$ pred, Prediction} \\
				\hline
			\end{tabular}
		\end{center}
		\caption{\label{tab:rule:sample} Example of a disambiguated sample in CBT-NE dataset with the proposed rule-based approach.}
	\end{table}
	
	\section{Attention Study}
	\label{sec:apx:att}
	
\begin{figure}[ht]
		\centering
		\includegraphics[width=\textwidth]{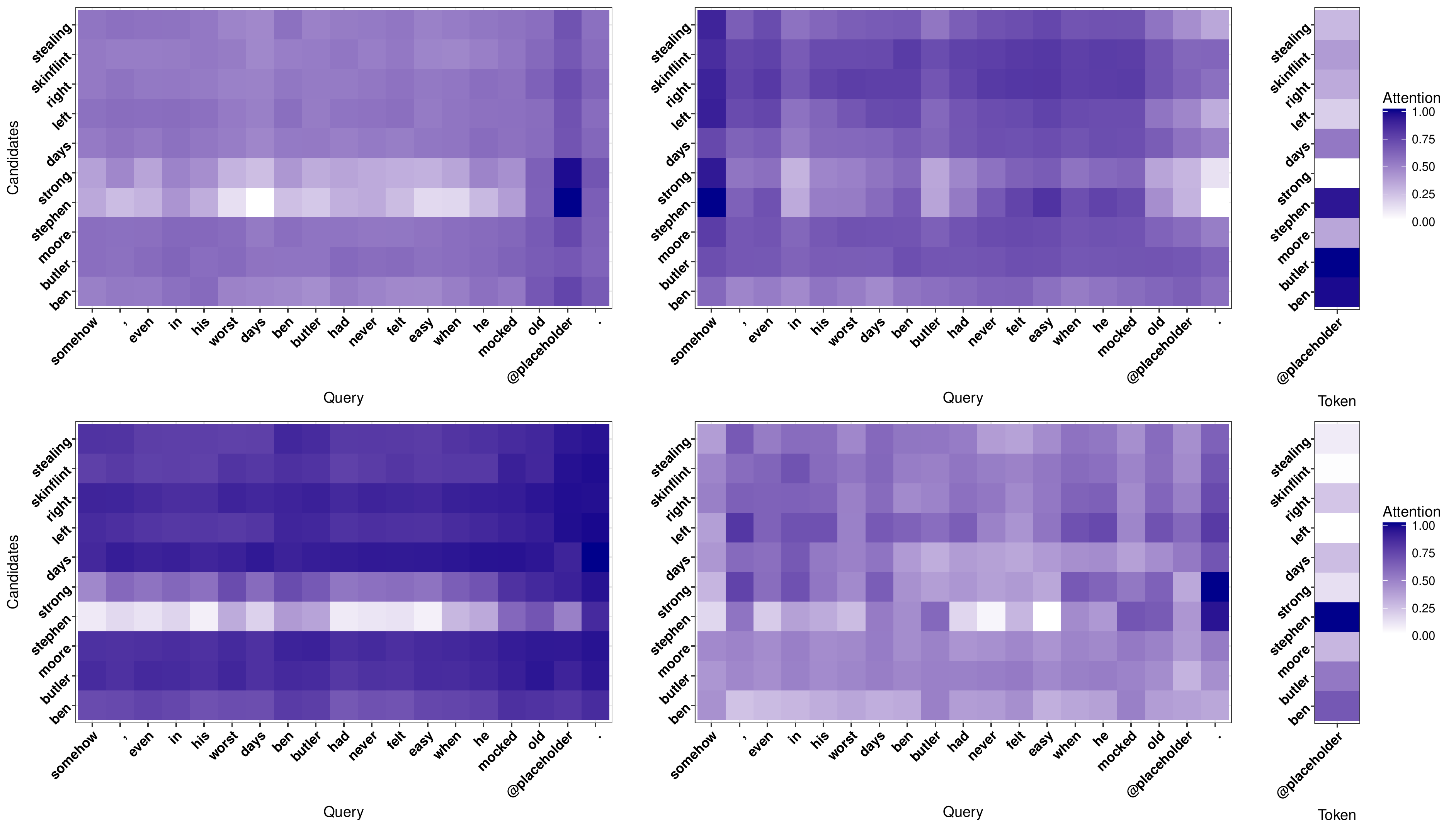}
		\caption{Layer-wise normalized attention visualization of ``DGR'' (top) and ``DGR - (a) \& (b) \& (c)'' (bottom) for a sample from the CBT-NE test set. Darker color illustrates higher attention. Figures only show the aggregated attention of candidates. The gold answer is ``butler''. \label{fig:apx:attention1}}
	\end{figure}

\begin{figure}[ht]
		\centering
		\includegraphics[width=\textwidth]{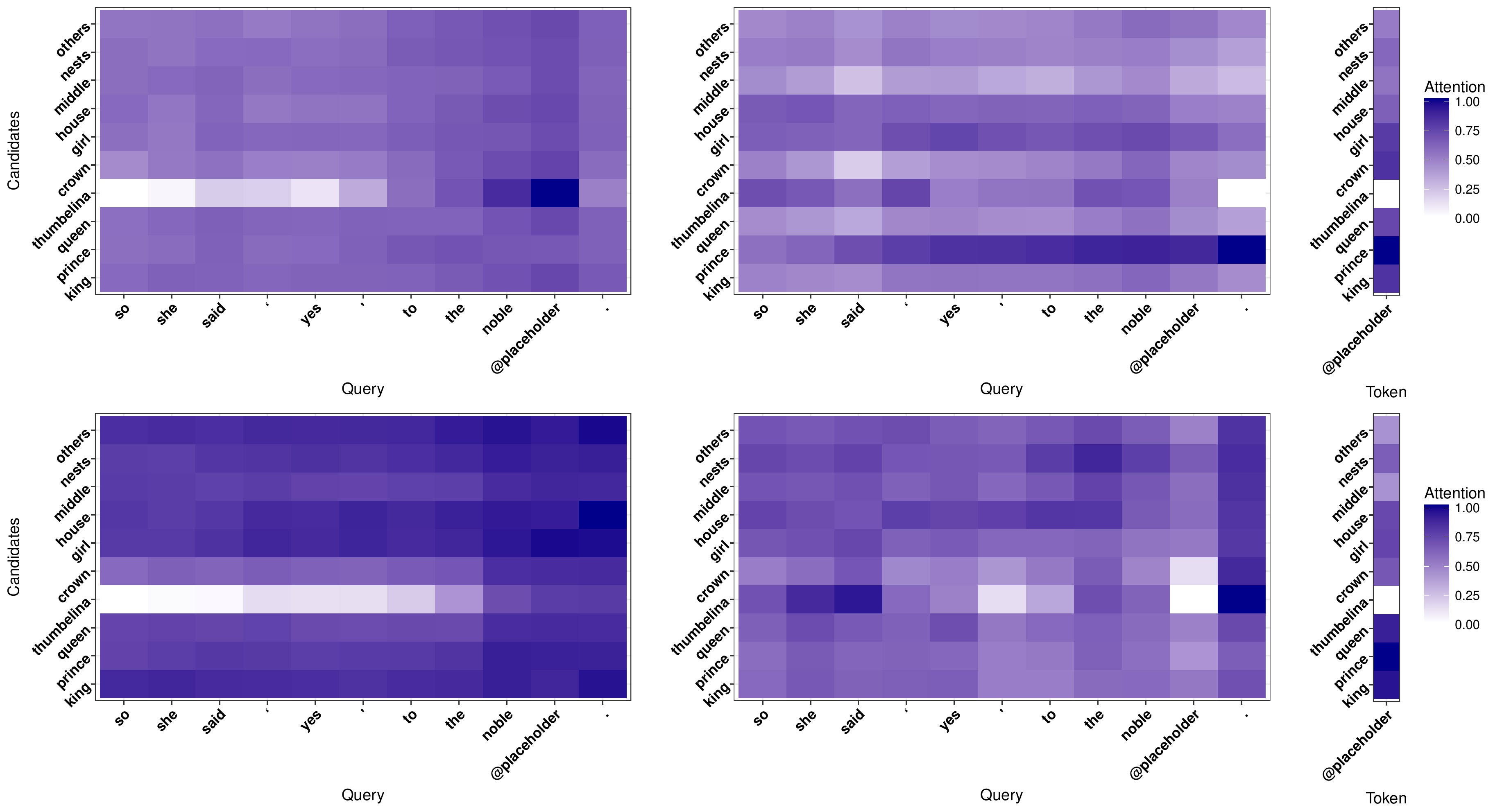}
		\caption{Layer-wise normalized attention visualization of ``DGR'' (top) and ``DGR - (a) \& (b) \& (c)'' (bottom) for a sample from the CBT-NE test set. Darker color illustrates higher attention. Figures only show the aggregated attention of candidates. The gold answer is ``prince''. \label{fig:apx:attention2}}
	\end{figure}

\begin{figure}[ht]
		\centering
		\includegraphics[width=\textwidth]{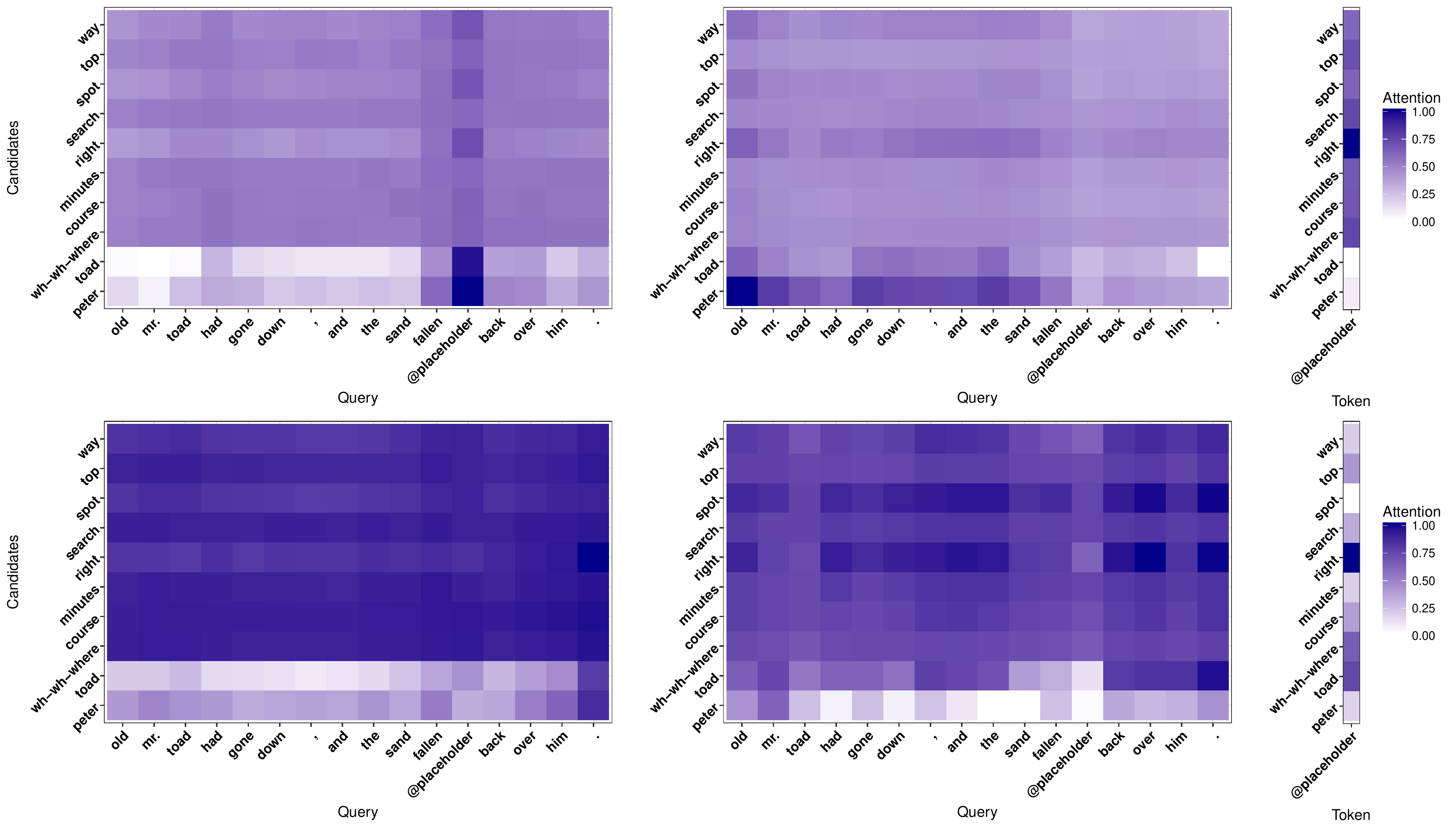}
		\caption{Layer-wise normalized attention visualization of ``DGR'' (top) and ``DGR - (a) \& (b) \& (c)'' (bottom) for a sample from the CBT-NE test set. Darker color illustrates higher attention. Figures only show the aggregated attention of candidates. The gold answer is ``right''. \label{fig:apx:attention3}}
	\end{figure}

\begin{figure}[ht]
		\centering
		\includegraphics[width=\textwidth]{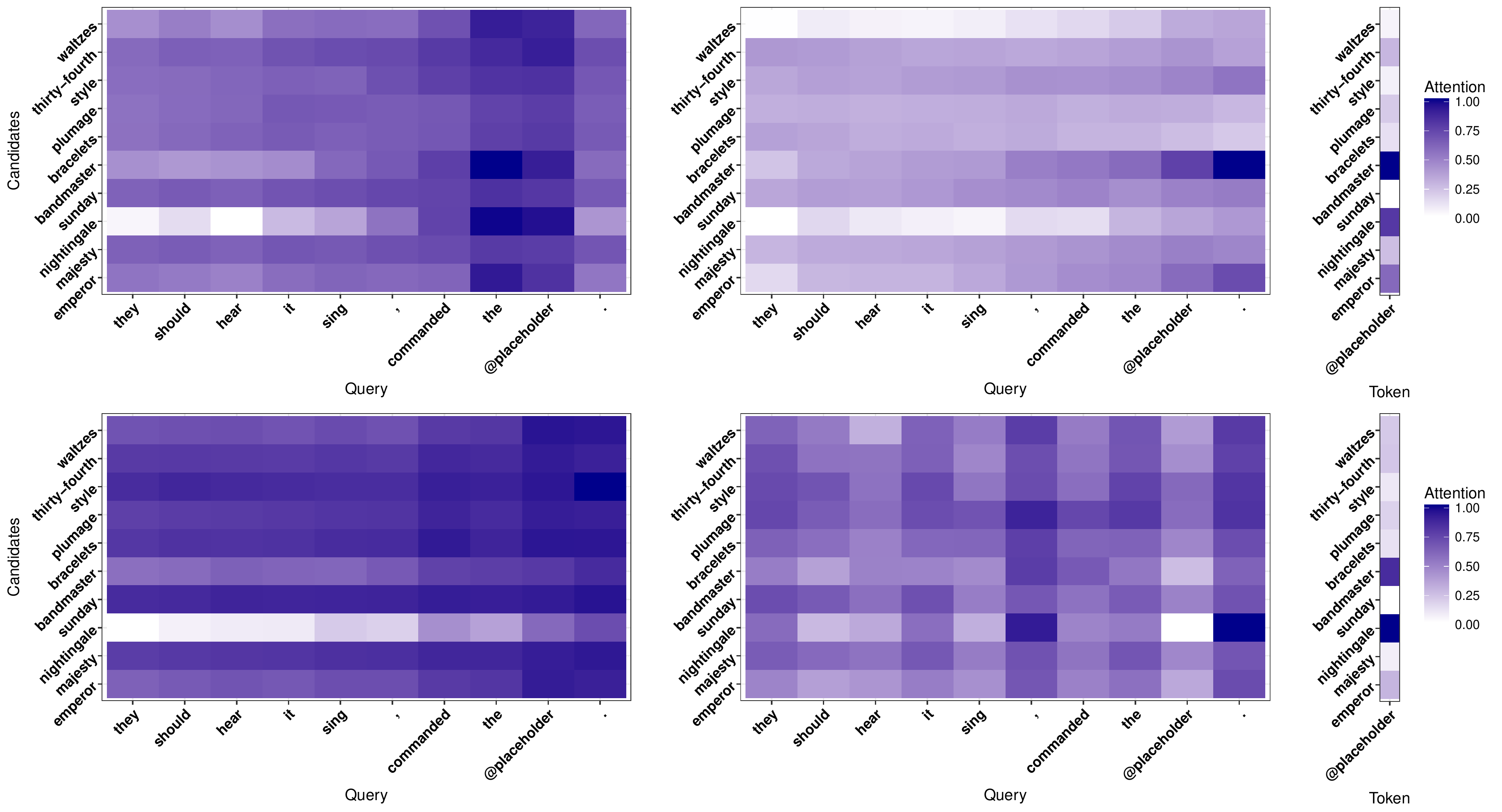}
		\caption{Layer-wise normalized attention visualization of ``DGR'' (top) and ``DGR - (a) \& (b) \& (c)'' (bottom) for a sample from the CBT-NE test set. Darker color illustrates higher attention. Figures only show the aggregated attention of candidates. The gold answer is ``bandmaster''. \label{fig:apx:attention5}}
	\end{figure}

\begin{figure}[ht]
		\centering
		\includegraphics[width=\textwidth]{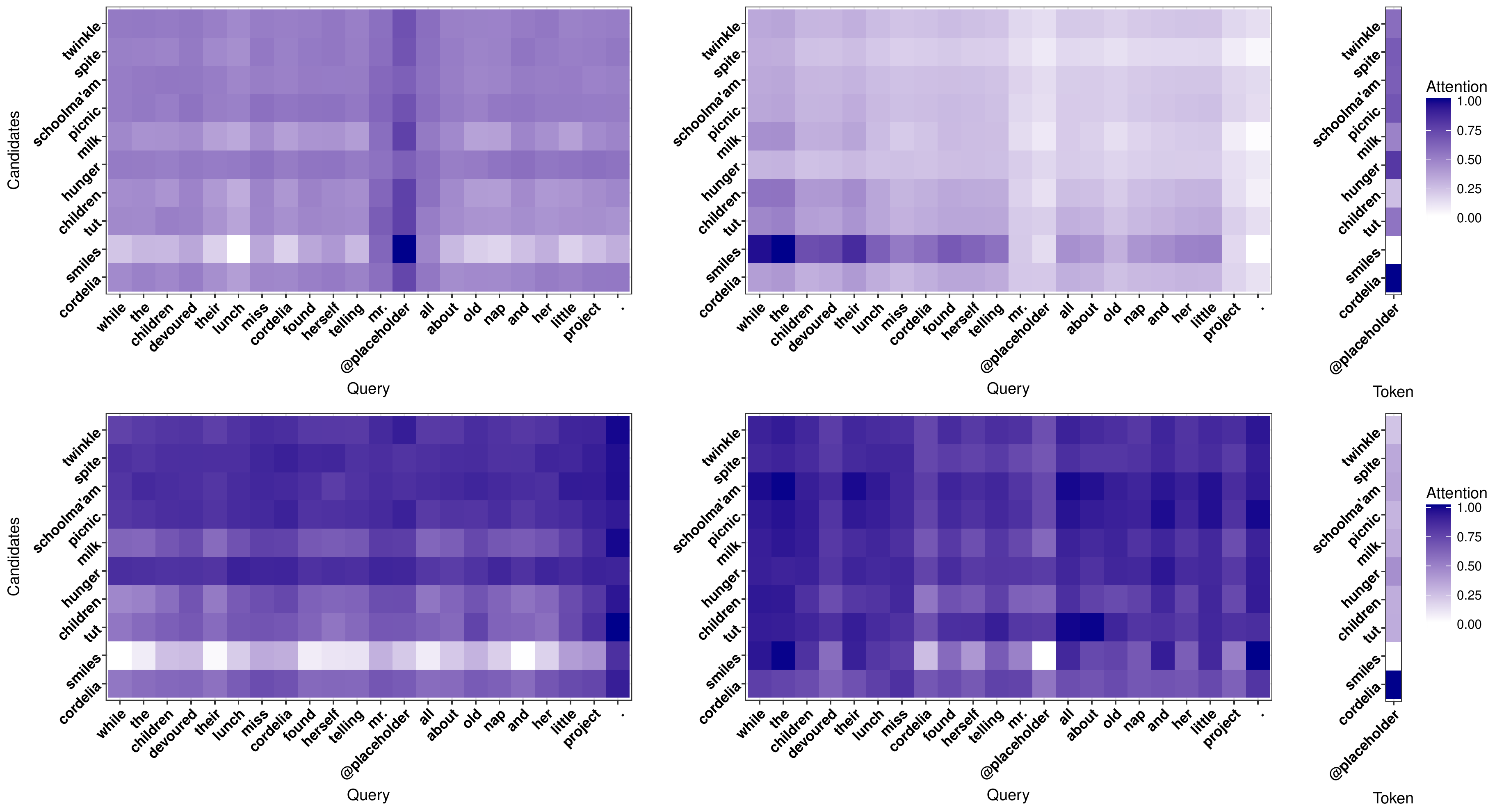}
		\caption{Layer-wise normalized attention visualization of ``DGR'' (top) and ``DGR - (a) \& (b) \& (c)'' (bottom) for a sample from the CBT-NE test set. Darker color illustrates higher attention. Figures only show the aggregated attention of candidates. The gold answer is ``cordelia''. \label{fig:apx:attention8}}
	\end{figure}

\begin{figure}[ht]
		\centering
		\includegraphics[width=\textwidth]{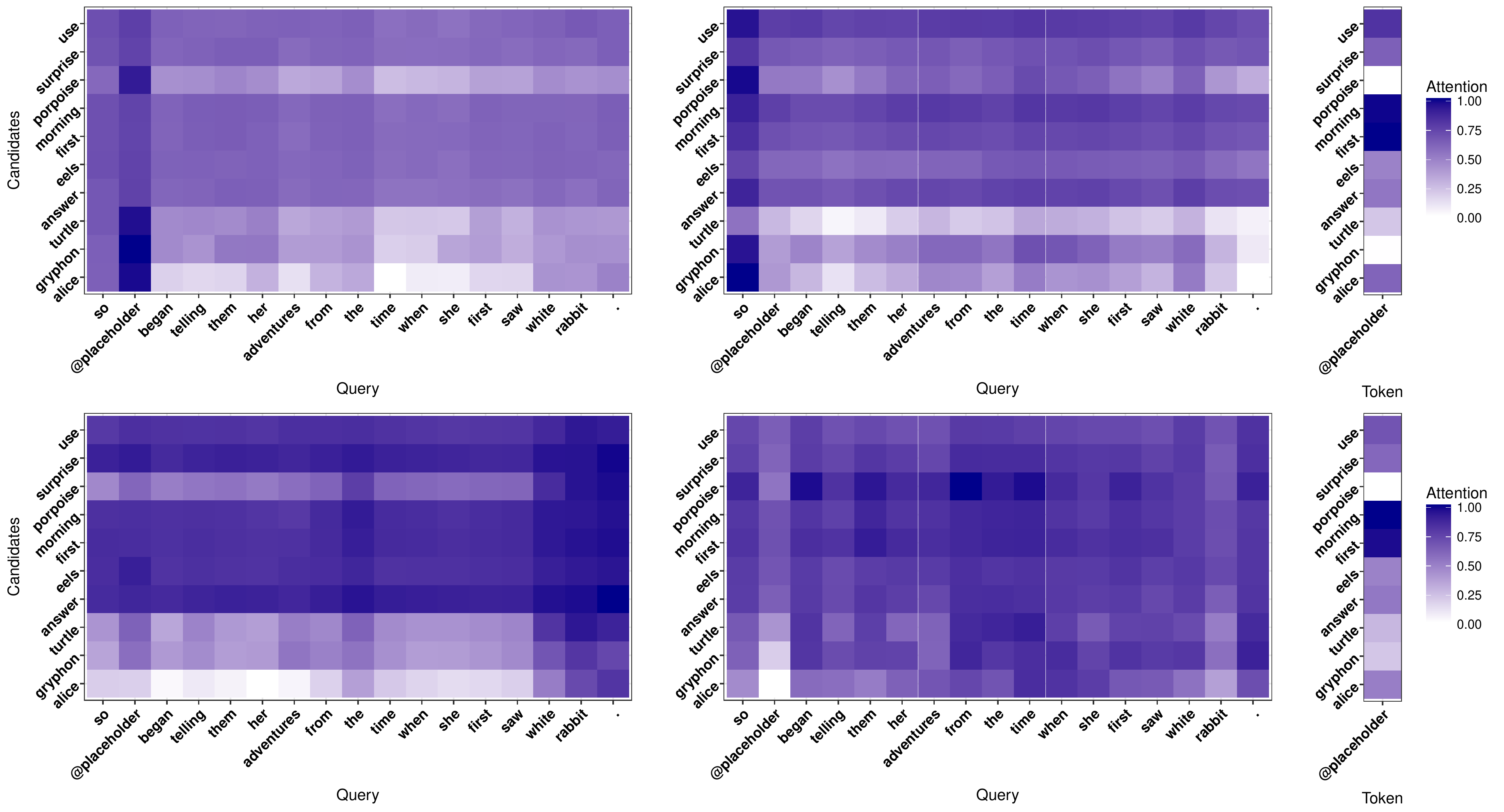}
		\caption{Layer-wise normalized attention visualization of ``DGR'' (top) and ``DGR - (a) \& (b) \& (c)'' (bottom) for a sample from the CBT-NE test set. Darker color illustrates higher attention. Figures only show the aggregated attention of candidates. The gold answer is ``first''. \label{fig:apx:attention9}}
	\end{figure}

\begin{figure}[ht]
		\centering
		\includegraphics[width=\textwidth]{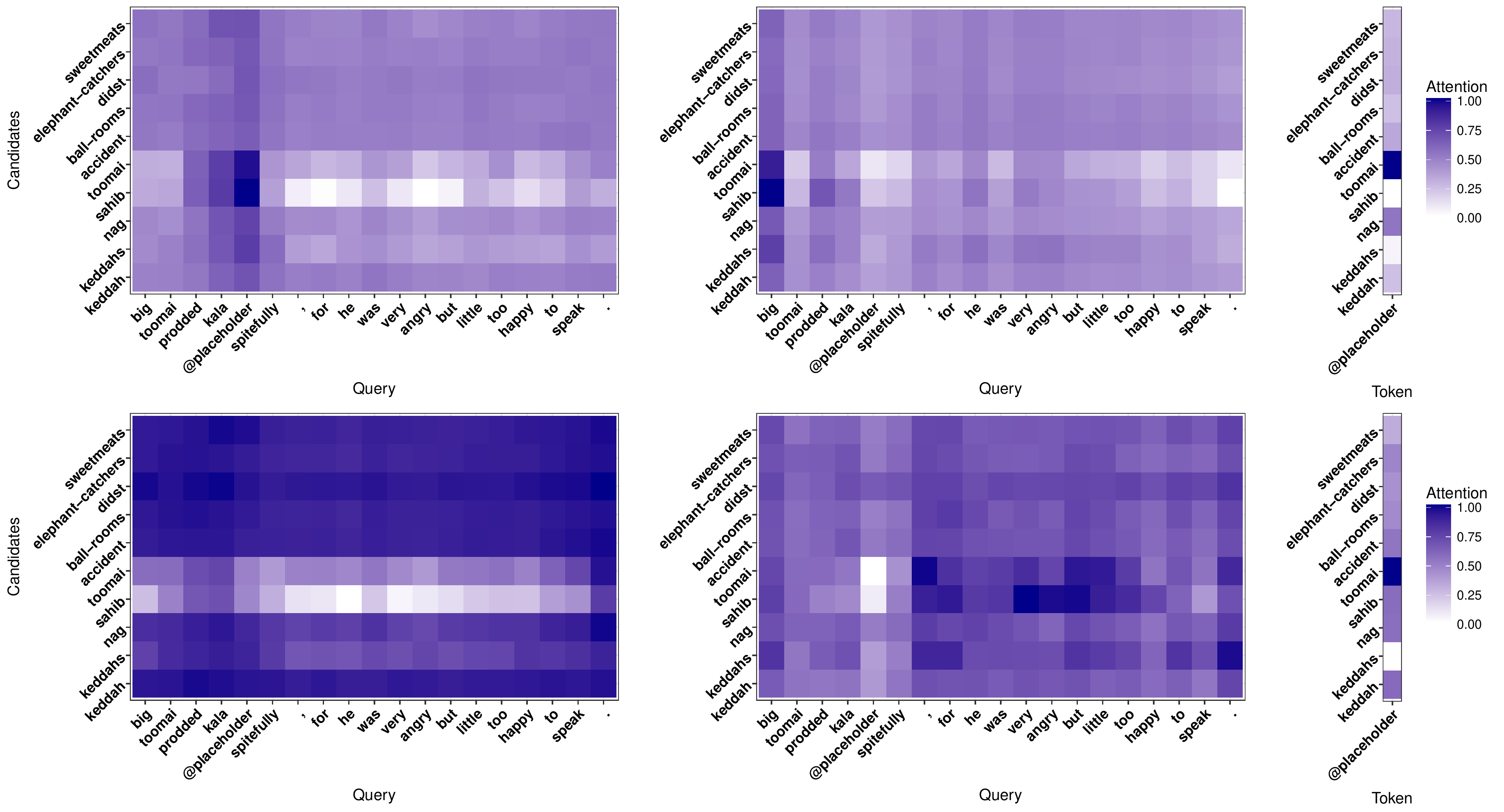}
		\caption{Layer-wise normalized attention visualization of ``DGR'' (top) and ``DGR - (a) \& (b) \& (c)'' (bottom) for a sample from the CBT-NE test set. Darker color illustrates higher attention. Figures only show the aggregated attention of candidates. The gold answer is ``toomai''. \label{fig:apx:attention10}}
	\end{figure}

\begin{figure}[ht]
		\centering
		\includegraphics[width=\textwidth]{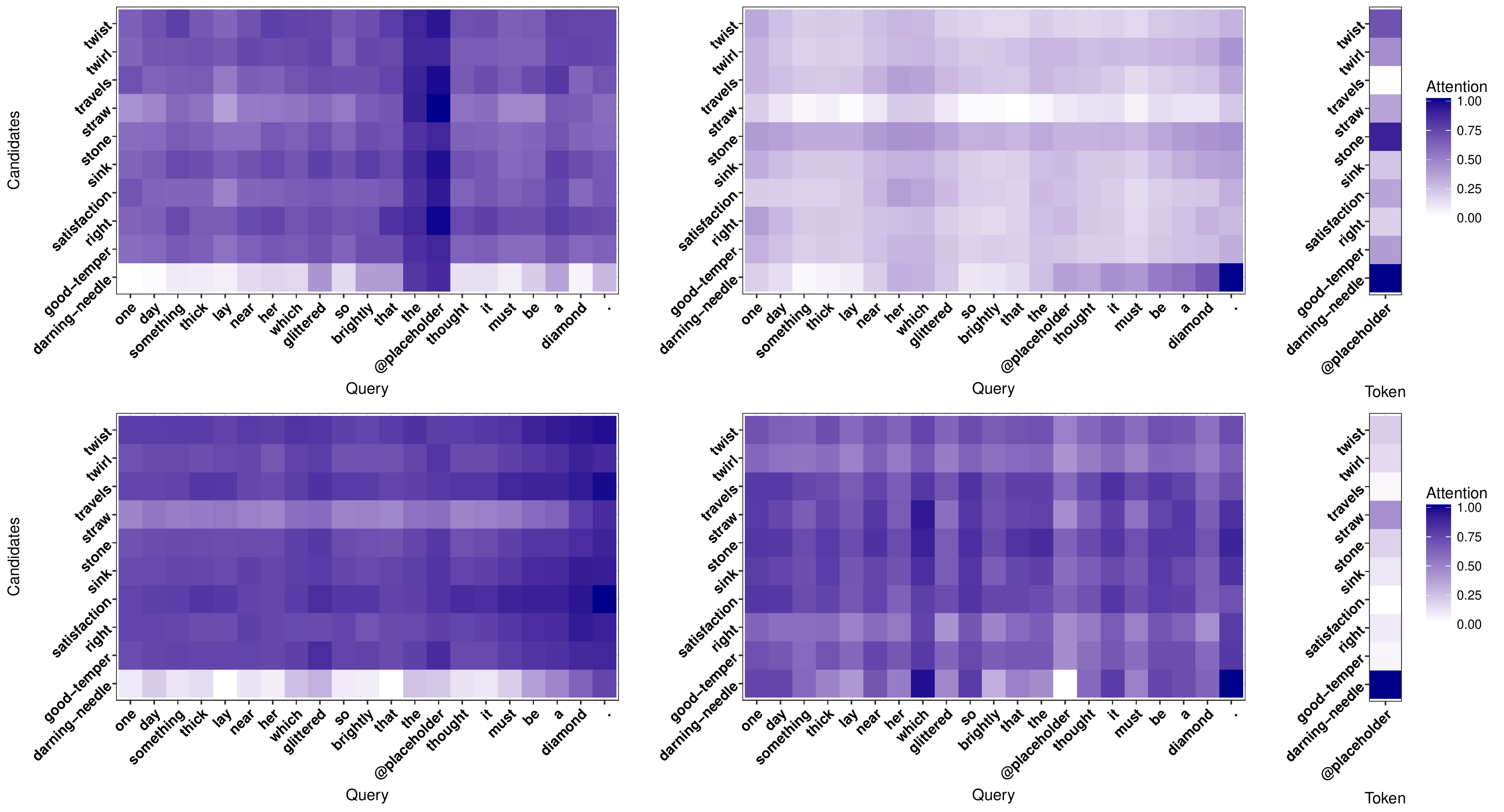}
		\caption{Layer-wise normalized attention visualization of ``DGR'' (top) and ``DGR - (a) \& (b) \& (c)'' (bottom) for a sample from the CBT-NE test set. Darker color illustrates higher attention. Figures only show the aggregated attention of candidates. The gold answer is ``darning-needle''. \label{fig:apx:attention11}}
	\end{figure}

	In this section, we show visualizations of 8 samples of layer-wise normalized attention (energy function, see Equation 3 in the main paper). Each column in Figures \ref{fig:apx:attention1}-\ref{fig:apx:attention11}, represents the same layer in ``DGR'' and ``DGR - (a) \& (b) \& (c)''. Also, each row is allocated to a specific model (Top: DGR, and Bottom: DGR - (a) \& (b) \& (c)).
	
\end{document}